# Automated Mouse Organ Segmentation:
# A Deep Learning Based Solution

Naveen Ashish
InferLink Corporation
2361 Rosecrans Ave, Suite 348
El Segundo, CA 90245
USA
nashish@inferlink.com

Mi-Youn Brusniak
Fred Hutch Cancer Research Center
1100 Fairview Ave N Seattle WA
98109
USA
mbrusnia@fredhutch.org

## ABSTRACT

The analysis of animal cross section images, such as cross sections of laboratory mice, is critical in assessing the effect of experimental drugs such as the biodistribution of candidate compounds in preclinical drug development stage. Tissue distribution of radiolabeled candidate therapeutic compounds can be quantified using techniques like *Quantitative Whole-Body Autoradiography* (QWBA). QWBA relies, among other aspects, on the accurate segmentation or identification of key organs of interest in the animal cross section image – such as the brain, spine, heart, liver and others. Currently, organs are identified manually in such mouse cross section images – a process that is labor intensive, time consuming and not robust, which leads to requiring a larger number of laboratory animals used. We present a deep learning based organ segmentation solution to this problem, using which we can achieve automated organ segmentation with high precision (dice coefficient in the 0.83-0.95 range depending on organ) for the key organs of interest.

## 1   INTRODUCTION

In the preclinical stage of drug discovery, the use of radiolabeled drug compounds offers the most efficient way to determine the tissue distribution of the compounds in laboratory animals. Quantitative Whole Body Autoradiography (QWBA), using phosphor-imaging technology has been routinely employed, in both developing new drugs as well as in addressing regulatory compliance needs. Radioactive signal from the compounds can be used to investigate whether the compounds reach the therapeutic target organ(s) and specific tissue(s), or whether the compounds show potential for off-target accumulation and toxicity for safety assessment. In any preclinical platform, enhancing the experimental robustness and reproducibility, while reducing cost and number of animals used, are important considerations. Thus, controlling technical variation in a QWBA platform is an important aspect in leading experiment design that can contribute to the reduction of animal use [1] and reproducibility. Solon and Kraus have pointed out several inconsistencies in QWBA studies for reproducibility [2]. One of these is identifying the region of the image(s) that are associated with radioactivity quantitative reported value. More specifically, lab scientists process black and white phosphor intensity images for determining organ boundaries for quantifying the radioactivity level in each of the different organs. Therefore, consistent and accurate identification of key organs in the digital images associated with these experiments is a key element in enhancing reproducibility and robustness in analyzing the effects of potential new drugs.

Due to the time consuming process, often times lab scientists identify and quantify only a few focused organs for a given study rather than acquiring quantitative data of all identifiable organs in a given section. As an example, Figure 1(a) illustrates a digital image of a mouse cross-section (rodents are typically used in such experiments), with all identifiable key organs. Figure 1 (b) illustrates the manual quantification of organs for a given specific study, with the brain, lung and kidney illustrated in this case. It is truly desirable to develop automated solutions that provide accurate, scalable and reproducible organ segmentation. This will reduce variance and collect more comprehensive data which in turn, we hope to reduce number of animal use and cost. This work describes an automated animal organ segmentation solution, developed in the context of mice and based on *deep learning*, to this problem.

Significant advancements have been made in the computer vision and automated image analysis fields in recent years, particularly powered by deep learning [3]. From recognizing cats and dogs to highly accurate classification and recognition of MRI or digital pathology images, deep learning is proving to be an effective and scalable technology for complex medical image analysis tasks. Capabilities such as automated segmentation of organs in animal images, powered by technologies such as deep learning, can significantly optimize the processes for drug discovery today.

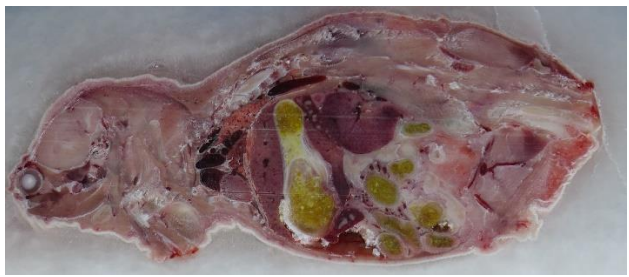

(a) Original mouse cross section

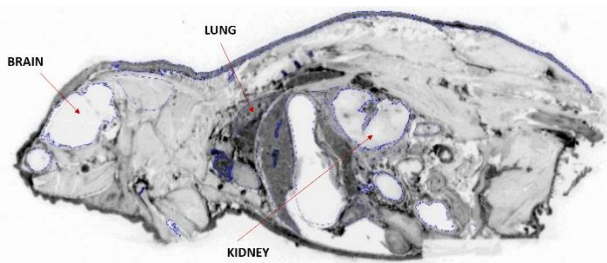

(b) Manual organ identification in black and white phosphor intensity images

**Figure 1: QWBA mouse cross sections**

We present an approach and implemented software pipeline for the automated segmentation of key organs in mice cross section images, taken as part of QWBA. Our approach is based on deep learning algorithms, but which are applied after some image processing operations on the images to be segmented. For the first phase of the work, our focus is on five key organs or image segments namely the (mouse) (i) brain, (ii) liver, (iii) tumor, (iv) kidney, and (v) spine.

Our solution and system have been developed and evaluated in the context of mouse organ segmentation, with mouse cross section images from the Optides research group at the Fred Hutchinson Cancer Research Center. The approach and system are, however, more generally applicable to organ segmentation in biomedical research.

## 1.1 Background: Autoradiography

Whole-body autoradiography (WBA) is an imaging process that determines the in situ localization of radiolabeled xenobiotics in laboratory animals [4]. This technique involves the dosing of animals, typically rodents, with radio labeled (typically 14C and 3H) compounds. Animals are euthanized at specified time points. The carcasses are snap-frozen, embedded in a frozen carboxymethylcellulose matrix, and cryosectioned. Whole-body cross sections that are 20 –50 microns thick, are obtained from different levels through the carcass and include representative samples of all major tissues. These are evaluated qualitatively and/or quantitatively for radioactive content using autoradiography and/or autoradioluminographical techniques. Autoradiography refers to the original technique of exposing the whole-body sections to X-ray film, which produces a photographic image. Phosphor imaging produces digital images of radioactivity distributed within the tissues of whole-body sections.

## 1.2 Medical Image Analysis: Related Work

Image processing and computing techniques based on shape models or an atlas have been developed since the nineties [5] [6]. In recent years, there has been significant development of machine-learning and particularly deep learning based approaches and technologies for image analysis. The segmentation problem has received attention as well, for instance approaches based on deep learning have been developed for tasks such as lung segmentation [7], the segmentation of organs in human anatomical 2D and 3D images such as from CT scans [8], the segmentation of particular organs (in human CT scans) such as pancreas segmentation [9], kidney segmentation [10], liver tumor segmentation [11] and also multi-organ segmentation [12]. Segmentation approaches are also being developed in many key medical image analysis driven applications other than organ segmentation, for instance the segmentation of tumors in brain images [13], lung segmentation in X-ray images [7], object detection in opthalmology images for diabetic retinopathy detection [14], and cell membrane and nuclei segmentation for breast cancer detection [15]. Automated segmentation has also been developed for non-biomedical domains, such as object segmentation in satellite imagery, computer vision for autonomous vehicles, and other applications [16].

A common technique for organ (or other object) segmentation is that of employing atlas-based or statistical shape models. The atlas-based segmentation relies on the existence of a reference image i.e., the atlas in which objects of interest have been manually segmented [17]. To segment an object in a new image, a transformation that registers the atlas to the patient image (establishing a point-to-point spatial correspondence) has to be computed. This transformation is what is then employed to segment out the desired object(s) in the new image. In the domain of organ segmentation in mouse cross-section images however, there are limiting factors namely 1) the data in terms of number of images per lab experiment is very small – about 50 cross-section images per experiment, 2) there is no standardized representation of such images in the domain – as it is though for other image domains such as (breast or other) cancer MRI images as an examples, and 3) there are no integrated archives or clearinghouse efforts for access to data from multiple groups. Certainly, elements from approaches such utilizing region-wise local atlas selection strategy [17], dictionary learning techniques [18] or leveraging inter-organ spatial relations [19], are applicable to our problem and have been employed in the approach we describe shortly.

There are machine learning based approaches for image classification, including pixel level classification [16]. We evaluated pixel level classification, taking the pixel location and color (R, G, B) as features and employing classifiers such as SVM and Random Forests to classify each pixel into one of the (5) organs or none. We were however, unable to achieve classification accuracy beyond 72 % precision with this approach. One obvious limitation is that organ or object level features, such as the organ shape or orientation are not taken into account at the pixel level. Deep learning on other hand, is driven off *unsupervised feature learning* [20] and the algorithms can better factor in aggregate information such as the features of pixels in the neighborhood of an individual pixel. Deep learning has been applied to many biomedical image analysis tasks as mentioned above. For segmentation, most of the deep learning based approaches take a pixel level segmentation approach, where for a (2D) image the objective is to determine a 2D matrix or "mask" that identifies element (pixel) by element its inclusion/not in the object to be segmented. Our evaluation of existing deep learning solutions for tasks such as lung segmentation [21] on our use case however, showed rather low segmentation accuracy. This is for the cases of both segmenting multiple organs within the entire mouse image as well as segmenting individual organs within smaller image zones (boxes) containing the organ. Recently, more comprehensive platforms for medical image analysis have emerged such as NiftyNet [22]. Such platforms provide a framework for integrated image data preparation, and analysis using machine (deep) learning which is useful but the aspect of determining which deep learning models or layers are effective for our problem remains open.

The reason existing deep learning classifiers for other medical segmentation tasks are not effective for our problem is that the mice cross section images have more complexity and ambiguity regards objects' location and boundary precision. Tasks like lung segmentation are in the context of black and white images, where there is a single object to be recognized (the lungs), and where the object has a fairly consistent shape, size and orientation across multiple images. The mouse cross section images on the other hand have complex colors and hues, and the objects to be detected have a relatively high variation of shape, size, and orientation in a collection. Also, organ boundaries are less well defined.

In our approach, we leverage knowledge about the representation of particular organs in images to convert the organ segmentation to a (simpler) shape detection problem. Our approach utilizes both image processing as well as machine and deep learning. We have assembled and evaluated custom deep learning classifiers, and have also explored transfer learning from pre-trained image recognition models including U-Net [23] for image segmentation and other models such as AlexNet [24], GoogleNet [25] and VGG16 [26] for classification for shape recognition. The next section is a detailed description of our solution. This is followed by experimental results and a conclusion discussion.

## 2 METHODS

Figure 2 illustrates the key steps and associated modules of the software pipeline that we have developed for this task, and that we refer to as the *organ segmentation pipeline*.

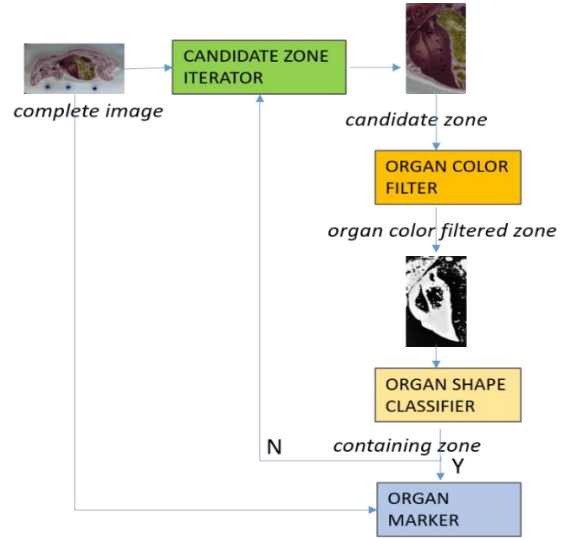

**Figure 2: Organ segmentation pipeline**

The pipeline functionality is to identify i.e., segment a specified organ in a complete mouse cross section image. As Figure 2 illustrates, the process starts with a complete mouse image, with then restricting the organ search to an (organ dependent) sub area within the complete mouse image, identifying a *bounding box* compactly containing the organ in this area, and finally segmenting the actual organ within this bounding box.

### 2.1 Data Preparation

The original images (from the experiments) are in JPEG format. Some data preparation (not shown in pipeline) involved (i) scaling the images to a uniform 2000 (W) X 1000 (H) pixels 2D size and (ii) normalizing the color intensity across images in a set. This was done with some custom code and utilities from the Python OpenCV [27] library. Ground truth data was created by manual annotation of organs, by three annotators.

### 2.2 Determining Organ Bounding Box

The *organ bounding box* is a rectangle that completely and compactly contains a specified organ. For instance in Figure 3, the red rectangles represent bounding boxes for the mouse brain and liver, respectively. The bounding box dimensions, for each organ, are fixed and determined from organ size statistics across multiple mouse images. body and thus the image. This sub-area is referred to as the organ *plausible region*.

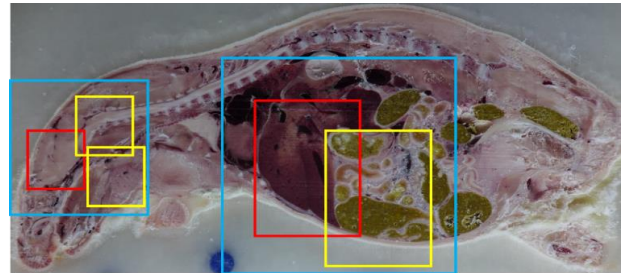

**Figure : Organ regions**

In Figure 3, the blue rectangles represent the plausible regions for the brain and the liver. We search for the bounding box for an organ in a sub-area of the complete mouse image, where the sub-area is determined by our knowledge of placement of organs in the mouse. The plausible region is identified by coordinates of the top left corner of the rectangle, where the top left corner of the entire mouse image is the origin (0, 0). For each organ, we determined the mean and standard deviation of the identifying (top left) corner of the organ bounding box across multiple mouse images. The range of this mean +- 3 standard deviations defines the plausible region for the organ. Table 1 provides the bounding box size and the plausible regions for the key organs, in a coordinate system where a mouse cross section image has dimensions 2000 (W) X 1000 (H) and the origin (0,0) at the image top left corner. Values are rounded to nearest tens.

**Table 1: Bounding boxes and plausible regions**

| Organ | Bounding Box size | Plausible region |
|---|---|---|
| Brain | 400 X 400 | [0, 400] to [120, 630] |
| Heart | 100 X 100 | [800, 430] to [990, 1000] |
| Liver | 300 X 800 | [1010, 400] to [1400, 710] |
| Kidney | 400 X 400 | [1200, 190] to [1500, 500] |
| Spine | 600 X 200 | [100, 50] to [400, 400] |

Figure 4 illustrates the actual process of generating candidate bounding boxes. For a new mouse image and organ to be identified in that image, we first retrieve the plausible region for the organ – that has been identified earlier from image statistics. We generate candidate bounding boxes, using a simple left-to-right and top-to-bottom exhaustive generation of *candidate* bounding boxes in the plausible region.

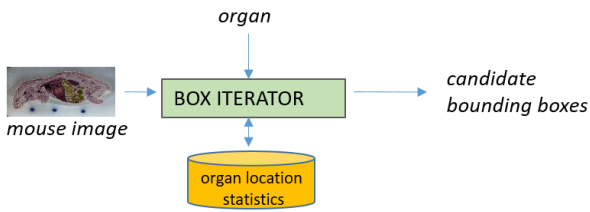

**Figure 4: Candidate bounding box iterator**

In Figure 3, the yellow boxes within the plausible region specified by the blue rectangle towards the left represent candidate bounding boxes for the brain , and the yellow box within the center blue rectangle (plausible region) represents a candidate bounding box for the liver. The (two) red boxes (left and center in Figure 3) represent true bounding boxes for the brain and liver, respectively.

## 2.3 Organ Color Filtering

For any candidate bounding box for an organ, we want to determine if it is a true bounding box or not. As mentioned earlier, our attempts to classify candidate bounding boxes as-is, using feature driven machine learning as well as deep learning, were unsuccessful in obtaining even a moderately accurate classification. Leveraging the color cohesiveness and consistency in individual organs, we translate the true bounding box determination problem to a *shape recognition* problem.

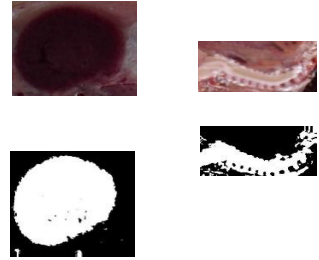

**Figure 5: Organ shapes**

Figure 5 illustrates the kidney and spine bounding box images with the original images on top and corresponding organ shapes below. Recognizing the shape (only) images is a relatively more tractable problem than recognizing an original image that has much higher heterogeneity of color across pixels. This is validated by our evaluation of deep learning classifiers applied to shape recognition, which achieved very high accuracy shape recognition accuracy. The questions then is – how do we transform an organ bounding box to a corresponding organ shape image ? This is done by filtering the (candidate) bounding box colors to retain o*nly* the colors that are among the category of colors for the organ. We employ a machine learning classifier determine the organ color category. The category of colors across organs are not always unique, for instance the liver and kidney share common colors. Table 2 lists the distinct categories we have categories the 5 organ colors into. The color category determination classifier itself is an SVM based classifier, trained on the R, G, B values of pixels of the organs. As Figure 6 illustrates, the organ color filter takes a candidate bounding box, employs a (pre-trained) organ color classifier to select and retain only the organ colors – as shown for the spine and kidney in this example.

**Table 2: Color categories**

| Color Category | Organs(s) |
|---|---|
| CAT1 | Brain |
| CAT2 | Spine |
| CAT3 | Heart |
| CAT4 | Liver, Kidney |

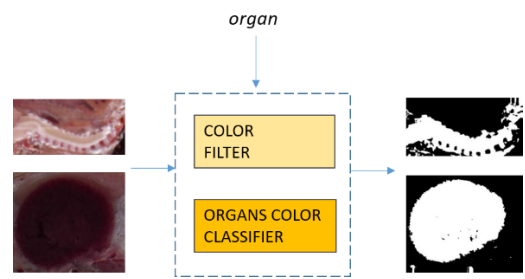

**Figure 6: Organ color filtering**

## 2.4 Organ Shape Recognition

As illustrated in Figure 7, we need to be able to distinguish the kidney or spine shape images from the *non* kidney or spine images (respectively) on the right. The organ shape image must contain the entire organ and not just have a partial representation.

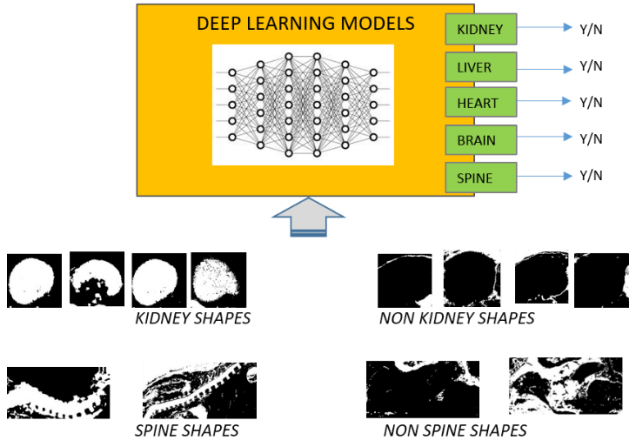

**Figure 7: Organ shape classifier**

We have found deep learning classifiers, based on our own models as well as pre-trained models, to be very effective in distinguishing between organ shape and non shape images, treating this as a classification task. The most popular architectures for image analysis, using machine or deep learning in general, are *convolutional neural networks* (CNNs) [28]. These CNNs are a fundamental element of the deep learning models we have composed, and are (also) a fundamental ingredient of the pre-trained models we have evaluated. We thus, first, provide a brief overview of CNNs.

### 2.3.1 Convolutional Neural Networks

Convolutional neural networks (CNNs) are a class of deep neural networks that have been applied with particular success to the image and video recognition domain. CNNs are basically deep, feed-forward artificial neural networks that use a variation of multilayer perceptrons (MLPs) designed to require minimal preprocessing. The CNN design is neurologically inspired in that the connectivity pattern between neurons resembles the organization of the animal visual cortex [20]. CNNs, like any deep learning network in general, require little or hand crafting for features for a machine learning task. The "convolution" in CNN is essentially an operator to extract features from an image, in the case of a 2D image it can be thought of as an operator that extracts features from the image by considering small squares of the input pixels where spatial relationships are preserved. CNNs usually work with local or global *pooling layers* which are essentially down sampling layers to reduce the dimensionality of the feature map. Pooling layers combine the outputs of neuron clusters at one layer into a single neuron in the next layer. For example, *max pooling* uses the maximum value from each of a cluster of neurons at the prior layer. Another example is *average pooling*, which uses the average value from each of a cluster of neurons at the prior layer. Finally, *fully connected* layers connect every neuron in one layer to every neuron in another layer. It is in principle the same as the traditional multi-layer perceptron neural network. Figure 7 illustrates a simple deep learning network for image classification with alternating convolution and poling layers, followed by fully connected layers leading into the final image classification.

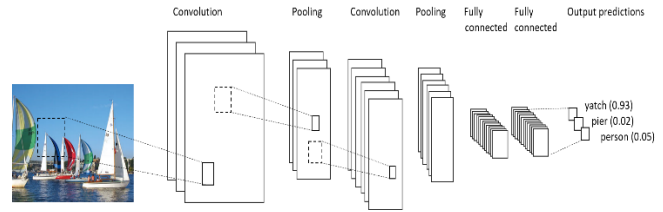

**Figure 7: Convolution and pooling layers**

### 2.3.2 Model for Shape Recognition

For the model we have composed, we have used alternating convolution and max pooling layers, followed by some dense layers and including a drop out layer. This is illustrated in Table 3, and Table 4 shows the corresponding Keras instructions.

**. Table 3: Deep learning model**

| Layer | Input | Filter | Output |
|---|---|---|---|
| $Conv_1$ | 2000 X 1000 | 3 X 3 | 2000 X 1000 |
| $Pooling_1$ | 2000 X 1000 | 2 X 2 | 1000 X 500 |
| $Conv_2$ | 1000 X 500 | 3 X 3 | 1000 X 500 |
| $Pooling_2$ | 1000 X 500 | 2 X 2 | 500 X 250 |
| $Conv_3$ | 500 X 250 | 3 X 3 | 500 X 250 |
| $Pooling_3$ | 500 X 250 | 2 X 2 | 250 X 125 |
| $Flatten_1$ | 250 X 125 | | 31250 X 1 |
| $Dense_1$ | 31250 X 1 | | 31250 X 1 |
| $Dense_2$ | 31250 X 1 | | 31250 X 1 |
| $Dropout_1$ | 31250 X 1 | | 31250 X 1 |
| $Dense_3$ | 31250 X 1 | | 6 X 1 |

**Table 4: Deep learning model in Keras/TensorFlow**

```
model = Sequential()
model.add(Conv2D(32,(3,3)))
input_shape=input_shape))
model.add(Activation('relu'))
model.add(MaxPooling2D(pool_size=(2, 2)))
model.add(Conv2D(32, (3, 3)))
model.add(Activation('relu'))
model.add(MaxPooling2D(pool_size=(2, 2)))
model.add(Conv2D(64, (3, 3)))
model.add(Activation('relu'))
model.add(MaxPooling2D(pool_size=(2, 2)))
model.add(Conv2D(64, (3, 3)))
model.add(Activation('relu'))
model.add(MaxPooling2D(pool_size=(2, 2)))
model.add(Conv2D(64, (3, 3)))
model.add(Activation('relu'))
model.add(MaxPooling2D(pool_size=(2, 2)))
model.add(Flatten())
model.add(Dense(128))
model.add(Activation('relu'))
model.add(Dense(64))
model.add(Activation('relu'))
model.add(Dropout(0.5))
model.add(Dense(1))
model.add(Activation('softmax'))
```

*2.3.3 Pre-trained Models*

General architectures for deep learning, manifested as pre-trained models, have seen significant traction in recent years. The power of many of these pre-trained models comes from their being trained on large collections of data – images in this case. For classification in particular, some of the widely used models include ImageNet, LeNet, AlexNet, GoogleNet and VGG16. All of these models have CNNs as a fundamental component. For segmentation, U-Net [23] is a prominent architecture and model and is also based on CNNs. As part of our experiments we evaluated both 1) U-Net for segmenting organs in the mouse images – the complete mouse image as well as parts of it, and 2) AlexNet, GoogleNet and VGG16 for their effectiveness in shape recognition.

## 2.5 Software Implementation

Our solution has been implemented as a functioning organ segmentation software pipeline. We have used Python [28] as the primary programing language, including relevant Python libraries such as Pandas [29], Numpy [30]. We have used the OpenCV image library and also ImageJ [31] for image processing. For machine learning we have used SciKit Learn [32], and TensorFlow [33] with Keras [34] for deep learning.

## 3 RESULTS

Our experimental evaluation evaluates the efficacy of our approach for organ segmentation, and also evaluates the applicability of other techniques and models to the problem. We have reported results in terms of the *dice coefficient* [35], which is measure of the overlap of the actual and classified segment. In the QWBA organ segmentation problem however, we prefer a more conservative segmentation approach where ensuring that pixels classified for an organ are indeed part of the organ is relatively more important than classifying all the pixels for the organ. In other words, the organ identification *precision* is more important than recall and we also report classification accuracy by precision and recall. Precision is the fraction [0-1] of *correctly* classified organ pixels to the pixels classified for the organ. We used a dataset of 100 original mouse cross-section image that was enhanced to 5000 images using OpenCV image data generation functions. A 50:50 split was made across training and test data.

## 3. 1 Segmentation Accuracy

Table 5 provides, by organ, the segmentation accuracy of our pipeline in terms of dice coefficient and precision. We also provide a comparison with 2 other approaches that we implemented and evaluated namely 1) deep learning based pixel by pixel classification and 2) regular SVM classifier based on position (coordinates) and color (R, G, B) features.

**Table 5: Organ segmentation accuracy**

| *Organ* | Dice-coefficient | Precision | Recall | F-score |
|---|---|---|---|---|
| Brain | 0.93 | 0.94 | 0.9 | 0.92 |
| Heart | 0.95 | 0.96 | 0.94 | 0.95 |
| Liver | 0.91 | 0.94 | 0.85 | 0.89 |
| Kidney | 0.94 | 0.96 | 0.89 | 0.92 |
| Spine | 0.83 | 0.87 | 0.79 | 0.83 |

In Table 6 we report on the segmentation accuracy (dice coefficient) of our initial approach evaluation namely (i) pixel level segmentation using deep learning and (ii) feature driven machine learning classification of each pixel for segmentation, using classifiers such as SVM.

**Table 6: Approach comparison**

| *Approach* *Organ* | Pipeline | Pixel level segmentation | Feature driven classification |
|---|---|---|---|
| Brain | 0.93 | < 0.1 | 0.61 |
| Heart | 0.95 | < 0.1 | 0.67 |
| Liver | 0.91 | < 0.1 | 0.53 |
| Kidney | 0.94 | < 0.1 | 0.74 |
| Spine | 0.83 | < 0.1 | 0.39 |

We also evaluated the U-Net convolutional network extensively, re-trained on our images and segmentation task. We attempted segmenting individual organs, in both whole mouse images, as well as within smaller organ plausible regions but achieved only very limited segmentation accuracy of dice coefficient < 0.1 in all cases. Our approach employs deep learning for shape classification, so we also evaluated various other shape classification models. Table 7 provides those results (dice coefficient values).

**Table 7: Shape recognition comparison**

|  | AlexNet | GoogleNet | VGG16 | Our model |
|---|---|---|---|---|
| Brain | 0.93 | 0.97 | 0.84 | 0.96 |
| Heart | 0.89 | 0.98 | 0.91 | 0.97 |
| Liver | 0.94 | 0.94 | 0.96 | 0.98 |
| Kidney | 0.92 | 0.96 | 0.90 | 0.99 |
| Spine | 0.89 | 0.87 | 0.73 | 0.94 |

### 3.2 Deep Learning Parameter Optimization

We also conducted parameter optimization. The variation segmentation accuracy in terms of are-under-curve (AUC), as a function of the number of (convolutional) layers and also training epochs is illustrated in Figure 8.

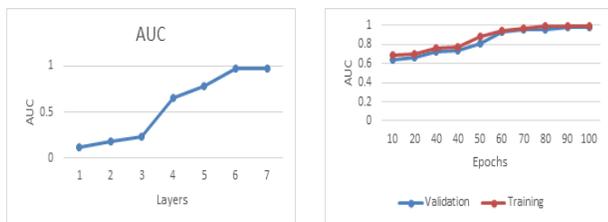

**Figure 8: Parameter optimization**

### 3.3 Throughput Performance

Model training takes an average of 254 seconds (70 epochs), and segmentation of an organ in a single image takes an average of 700ms. This is on a 4 node in-house computing cluster, running Ubuntu 14.04. The cluster is drawn from a larger, organization wide computing cluster currently equipped with 504 compute nodes, 2664 CPU cores and more than 24 TB of main memory (RAM). Each cluster node contains 1-6 TB local disk space which can be used for non-shared temporary data.

## 4 DISCUSSION

We have presented a solution for automated organ segmentation in the context of mouse cross section images, with an initial focus on five key organs. We are currently starting the evaluation of the applicability of the approach and system to all detectable organs to generate consist and comprehensive data acquisition from each experiment involving mouse cross section images. The long term goal of our work is to develop a comprehensive platform for biomedical image analysis that is motivated by use cases in cancer research for image analysis capabilities such as segmentation and classification. A more immediate step in this direction is to expand the current mouse organ segmentation capabilities of our software system, to other organ segmentation problems at our center such as human anatomical organ segmentation. We will also extend the capabilities to other image segmentation problems, such as segmenting micronuclei in fluorescent microscopy cell images and other use cases.

## ACKNOWLEDGEMENTS
This work was supported by Fred Hutchinson Cancer Research Center and philanthropic support from Project Violet for cancer cures drug discovery.

## REFERENCES

[1] J. I. Everitt, "The Future of Preclinical Animal Models in Pharmaceutical Discovery and Development: A Need to Bring In Cerebro to the In Vivo Discussion," *Toxicologic Pathology, 43,* pp. 70-77, 2014.

[2] E. Solon, "Autoradiography techniques and quantification of drug distribution," *Cell and Tissue Research 360(1),* pp. 87-107, 2015.

[3] G. Litkens, T. Kooi, B. Bejnordi, A. Setio, F. Ciompi, M. Ghafoorian, J. Laak, B. v. Ginneken and C. Sánchez, "A survey on deep learning in medical image analysis," *Medical Image Analysis, Vol 42,* pp. 60-88, 2017.

[4] A. McEwen and C. Henson, "Quantitative whole-body autoradiography: past, present and future," *Bioanalysis 7(5),* pp. 557-568, 2015.

[5] L. G. Shapiro and G. C. Stockman, Computer Vision., Prentice Hall., 2001.

[6] W. Freeman, P. Perona and B. Scholkopf, "Guest Editorial: Machine Learning for Computer Vision," *International Journal of Computer Vision. 77 (1),* 2008.

[7] Yu.Gordienko, P. Gang, J. Hui, W. Zeng, Yu.Kochura, O.Alienin, O. Rokovyi and S. Stirenko, "Deep Learning with Lung Segmentation and Bone Shadow Exclusion Techniques for Chest X-Ray Analysis of Lung Cancer," *CoRR,* p. abs/1712.07632 , 2018.

[8] Zheng, Y. &. Liu, D. &. Georgescu, B. &. Xu, D. &. Comaniciu and Dorin, "Deep Learning Based Automatic Segmentation of Pathological Kidney in CT: Local Versus Global Image Context," in *Advances in Computer Vision and Pattern Recognition*, 2017, pp. 241-255.

[9] H. Roth, L. Lu, A. Farag, H. Shin, J. Liu, E. Turkbey and R. Summer, "Multi-level deep convolutional networks for automated pancreas segmentation.," *Ronneberger, O., P.Fischer, & Brox, T. (2015). U-net: Convolutional networks for biomedical image segmentation. Medical Image Computing Computer Assisted Interventions,* p. 556–564, 2015.

[10] Thong, W., Kadoury, S., Pich´e, N., Pal and C. J., "Convolutional networks for kidney segmentation in contrast-enhanced CT scans," *Computer Methods in Biomechanics and Biomedical Engineering: Imaging & Visualization,* pp. 277-282, 2016.

[11] Vivanti, R., Ephrat, A., Joskowicz, L., Karaaslan, O., Lev-Cohain, N., Sosna and J., "Automatic liver tumor segmentation in followup CT studies using convolutional

neural networks," in *MICCAI: Patch-Based Methods in Medical Image Processing Workshop*, 2015.

[12] P. Hu, F. Wu, J. Peng, Y. Bao, F. Chen and D. Kong, "Automatic abdominal multi-organ segmentation using deep convolutional neural network and time-implicit level sets," *International Journal of Computer Assisted Radiology Surgery. 12(3),* pp. 399-411, 2017.

[13] Zhang, W., Li, R., Deng, H., Wang, L., Lin, W., Ji, S., Shen and D., "Deep convolutional neural networks for multi-modality isointense infant brain image segmentation," *NeuroImage 108,* p. 214–224, 2015.

[14] E. Rahimy, "Deep learning applications in ophthalmology," *Current Opinion in Ophthalmology. 29(3),* pp. 254-260, 2018.

[15] M. Saha and C. Chakraborty, "Her2Net: A Deep Framework for Semantic Segmentation and Classification of Cell Membranes and Nuclei in Breast Cancer Evaluation," *IEEE Transactions on Image Processing 27(5),* pp. 2189-2200, 2018.

[16] C. Steger, M. Ulrich and C. Wiedemann, Machine Vision Algorithms and Applications (2nd ed.), Weinheim: Wiley-VCH, 2018.

[17] V. Duay, N. Houhou and J.-P. Thiran, "ATLAS-BASED SEGMENTATION OF MEDICAL IMAGES LOCALLY CONSTRAINED BY LEVEL SETS," Signal Processing Institute (ITS) Ecole Polytechnique F´ed´erale de Lausanne (EPFL) TR 07/2005, 2005 .

[18] C. Bao, H. Ji, Y. Quan and Z. Shen, "Dictionary Learning for Sparse Coding: Algorithms and Convergence Analysis," *IEEE Transactions on Pattern Analysis and Machine Intelligence 38(7),* pp. 1356 - 1369, 2015.

[19] C. Jud, "Object segmentation by fitting statistical shape models : a Kernel-based approach with application to wisdom tooth segmentation from CBCT images," 2014. [Online]. Available: https://edoc.unibas.ch/34098/.

[20] I. Goodfellow, Y. Bengio and A. Courville, Deep Learning, MIT Press, 2016.

[21] K. Kuan, M. Ravaut, G. Manek, H. Chen, J. Lin, B. Nazir, C. Chen, T. C. Howe, Z. Zeng and V. Chandrasekhar, "Deep Learning for Lung Cancer Detection: Tackling the Kaggle Data Science Bowl 2017 Challenge," *CoRR,* p. abs/1705.09435 , 2017.

[22] E. Gibson, W. Li, C. Sudre, L. Fidon, D. Shakir, G. Wang, Z. Eaton-Rosen, R. Gray, T. Doel, Y. Hu, T. Whyntie, P. Nachev, M. Modat, D. Barratt, S. Ourselin, M. Cardoso and T. Vercauteren, "NiftyNet: a deep-learning platform for medical imaging," *Computer Methods and Programs in Biomedicine,* pp. 113-122, 2018.

[23] O. Ronneberger, P.Fischer and T. Brox, "U-net: Convolutional networks for biomedical image segmentation," *Medical Image Computing Computer Assisted Interventions,* pp. 234 - 241, 2015.

[24] A. Krizhevsky, I. Sutskever and G. E. Hinton, "ImageNet classification with deep convolutional neural networks," in *Proceedings of the 25th International Conference on Neural Information Processing Systems*, 2012.

[25] C. Szegedy, W. Liu, Y. Jia, P. Sermanet, S. Reed, D. Anguelov, D. Erhan, V. Vanhoucke and A. Rabinovich, "Going Deeper with Convolutions," in *Computer Vision and Pattern Recognition (CVPR)*, 2015.

[26] K. Simonyan and A. Zisserman, "Very Deep Convolutional Networks for Large-Scale Image Recognition," *CoRR,* p. abs/1409.1556, 2015.

[27] "Open Source Computer Vision Library (OpenCV)," 2018. [Online]. Available: https://opencv.org/.

[28] "Python programming language," 2018. [Online]. Available: http://www.python.org.

[29] "Pandas," 2018. [Online]. Available: https://pandas.pydata.org/.

[30] "NumPy," 2018. [Online]. Available: http://www.numpy.org/.

[31] "ImageJ: Image Processing and Analysis in Java," 2018. [Online]. Available: https://imagej.nih.gov/ij.

[32] "scikit-learn : Machine Learning in Python," 2018. [Online]. Available: http://scikit-learn.org.

[33] "TensorFlow," 2018. [Online]. Available: https://www.tensorflow.org/.

[34] "Keras," 2018. [Online]. Available: https://keras.io/.

[35] T. Sorensen, "A method of establishing groups of equal amplitude in plant sociology based on similarity of species and its application to analyses of the vegetation on Danish commons," *Kongelige Danske Videnskabernes Selskab 5(4),* p. 1034, 1948.

[36] Xing, F., Xie, Y., Yang and L., "An automatic learning-based framework for robust nucleus segmentation," *IEEE Transactions on Medical Imaging 35(2),* pp. 550-566, 2016.

[37] Song, Y., Zhang, L., Chen, S., Ni, D., Lei, B., Wang and T., "Accurate segmentation of cervical cytoplasm and nuclei based on multiscale convolutional network and graph partitioning," *IEEE Transactions of Biomedical Engineering,* p. 2421–2433, 2015.

[38] E. Shelhamer, J. Long and T. Darrell, "Fully Convolutional Networks for Semantic Segmentation," *IEEE Transactions on Pattern Analysis and Machine Intelligence 39(4),* pp. 640-651, 2017.

[39] O. Ronneberger, P. Fischer and T. Brox, "U-Net: Convolutional Networks for Biomedical Image Segmentation," in *International Conference on Medical Image Computing and Computer-Assisted Intervention*, 2015.

[40] W. Rawat and Z. Wang, "Deep Convolutional Neural Networks for Image Classification: A Comprehensive Review," *Neural Computation 29,* p. 2352–2449, 2017.

[41] M. P. McBee, O. A. Awan, A. T. Colucci, C. W. Ghobadi, N. Kadom, A. P. Kansagra, S. Tridandapani and W. F. Auffermann, "Deep Learning in Radiology," *Academic Radiology,* p. In press, 2018.

[42] Y. Li and L. Shen, "Skin Lesion Analysis towards Melanoma Detection Using Deep Learning Network," *Sensors 18(2),* 2018.

[43] Y. L, G. Y, W. Y, Y. J and C. P, "Segmentation of Fetal Left Ventricle in Echocardiographic Sequences Based on Dynamic Convolutional Neural Networks," *IEEE Transactions on Biomedical Engineering 64(8),* pp. 1886-1895, 2017.

[44] E. Gibson, W. Li, C. Sudre, L. Fidon, D. I. Shakir, G. Wang, Z. Eaton-Rosen, R. Gray, T. Doel, Y. Hu, T. Whyntie, P. Nachev, M. Modat, D. C. Barratt, S. Ourselin, M. J. Cardoso and T. Vercauteren, "NiftyNet: a deep-learning platform for medical imaging," *Computer Methods and Programs in Biomedicine,* pp. 113-122, 2018.

[45] A. Garcia-Garcia, S. Orts, S. Oprea, V. Villena-Martinez and J. G. Rodriguez, "A Review on Deep Learning Techniques Applied to Semantic Segmentation," *CoRR,* p. abs/1704.06857, 2017.

[46] N. Ashish and A. Patawari, "Data Dictionary Reader Code," 2017. [Online]. Available: https://github.com/nashish100/DDReading.

[47] R. R. Holgers, L. Lu, A. Farag, H.-C. Shin, J. Liu, E. B. Turkbey and R. M. Summers, "DeepOrgan: Multi-level Deep Convolutional Networks for Automated Pancreas Segmentation," in *International Conference on Medical Image Computing and Computer-Assisted Intervention MICCAI*, 2015.